\documentclass[10pt,twocolumn,letterpaper]{article}

\usepackage{iccv}
\usepackage{times}
\usepackage{epsfig}
\usepackage{graphicx}
\usepackage{amsmath}
\usepackage{amssymb}
\usepackage{color}
\usepackage{booktabs}
\usepackage{makecell}
\usepackage{multirow}
\usepackage{bbding}
\usepackage{pifont}
\usepackage{xcolor}
\usepackage{comment}
\usepackage{appendix}
\newcommand{\cmark}{\ding{51}}%
\newcommand{\xmark}{\ding{55}}%

\usepackage[pagebackref=true,breaklinks=true,letterpaper=true,colorlinks,bookmarks=false]{hyperref}

\iccvfinalcopy 

\def\iccvPaperID{2776} 

\ificcvfinal\fi
\newcommand{\yukang}[1]{{\color{brown} #1}}

\begin{document}

\title{IST-Net: Prior-free Category-level Pose Estimation with \\ Implicit Space Transformation}

\author{Jianhui Liu$^1$,~\, Yukang Chen$^2$,~\, Xiaoqing Ye$^{3}$,~\, Xiaojuan Qi$^{1 \dagger}$\\
$^1$The University of Hong Kong,~\, $^2$The Chinese University of Hong Kong,~\, $^3$Baidu\\
{\tt\small jhliu0212@gmail.com, xjqi@eee.hku.hk, yukangchen@cse.cuhk.edu.hk, yexiaoqing@baidu.com}
}

\maketitle
\ificcvfinal\thispagestyle{empty}\fi

\begin{abstract}

Category-level 6D pose estimation aims to predict the poses and sizes of unseen objects from a specific category. Thanks to prior deformation, which explicitly adapts a category-specific 3D prior (\textit{i.e.}, a 3D template) to a given object instance, prior-based methods attained great success and have become a major research stream. However, obtaining category-specific priors requires collecting a large amount of 3D models, which is labor-consuming and often not accessible in practice. This motivates us to investigate whether priors are necessary to make prior-based methods effective. Our empirical study shows that the 3D prior itself is not the credit to
the high performance. The keypoint actually is the explicit deformation process, which aligns camera and world coordinates supervised by world-space 3D models (also called canonical space). Inspired by these observations, we introduce a simple prior-free implicit space transformation network, namely \textbf{IST-Net}, to transform camera-space features to world-space counterparts and build correspondence between them in an implicit manner without relying on 3D priors. Besides, we design camera- and world-space enhancers to enrich the features with  pose-sensitive information and geometrical constraints, respectively.
Albeit simple, IST-Net achieves state-of-the-art performance based-on prior-free design, with top inference speed on the REAL275 benchmark. Our code and models are available at \url{https://github.com/CVMI-Lab/IST-Net}. 

\end{abstract}

\begin{figure}[t]
\begin{center}
   \includegraphics[width=1.0\linewidth]{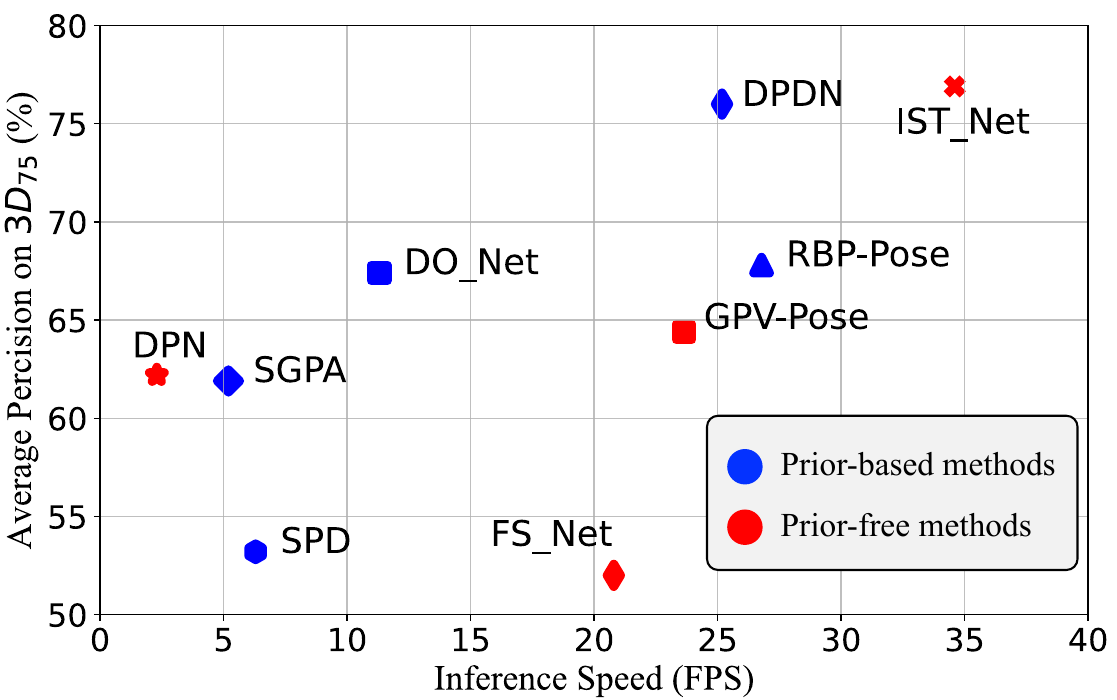}
   \caption{Comparison with competitive methods on REAL275 dateset. DPN refers to the DualPoseNet~\cite{dualpose-net}. All speeds are measured on a single RTX3090Ti GPU. We use blue/red to distinguish prior-based/free methods. IST-Net achieves top performance on $3D_{75}$ with the best inference speed.} 
   \label{fig:method-compare}
\end{center}
\vspace{-0.3cm}
\end{figure}

\section{Introduction}

Category-level pose estimation~\cite{wang2019normalized} draws great attention and plays an important role in practical applications, including 
robotic manipulation~\cite{deng2020self, mousavian20196, li2021simultaneous}, augmented reality~\cite{su2019deep}, and scene understanding~\cite{nie2020total3dunderstanding, zhang2021holistic}. 
Unlike instance-level pose estimation~\cite{ssd6d, labbe2020cosypose, li2018deepim, peng2019pvnet, he2020pvn3d, xiang2017posecnn}, which requires a 3D CAD model for each object instance, this task aims at exploiting category-specific information and thus can further generalize to unseen objects within given categories.

Recently, many methods~\cite{dpdn, sgpa, tian2020shape, CR-Net,fs-net, DBLP:journals/corr/abs-2001-09322, DBLP:journals/corr/abs-2008-08145} have been proposed for category-level pose estimation, which can be categorized into two groups: prior-free methods and prior-based methods. Prior-free methods~\cite{fs-net,dualpose-net,wang2019normalized, wang20206} mainly focus on designing network structures to fit the training data better.  
These methods are relatively simple but struggle to generalize to novel objects  and suffer from poor performance. 

To address this issue, prior-based methods ~\cite{dpdn,sgpa, tian2020shape, GPV-pose, lee2022uda, fu2022category} leverage category-specific 3D priors (templates) to guide pose estimation. They  adopt a prior-driven deformation module~\cite{tian2020shape} to  deform the prior for synthesizing the target object in world-space. And then, they formulate the pose estimation problem as a camera- and world-space correspondence learning  problem which explicitly aligns the coordinates~\cite{tian2020shape}. Although considerable progress has been attained with prior-based methods ~\cite{sgpa, zhang2022rbp, CR-Net}, the requirements of collecting a large amount of ground-truth 3D models of target objects for obtaining the 3D prior and supervising training the prior deformation module hinders their practical applicability. 

\begin{figure*}[t]
\begin{center}
   \includegraphics[width=1.0\linewidth]{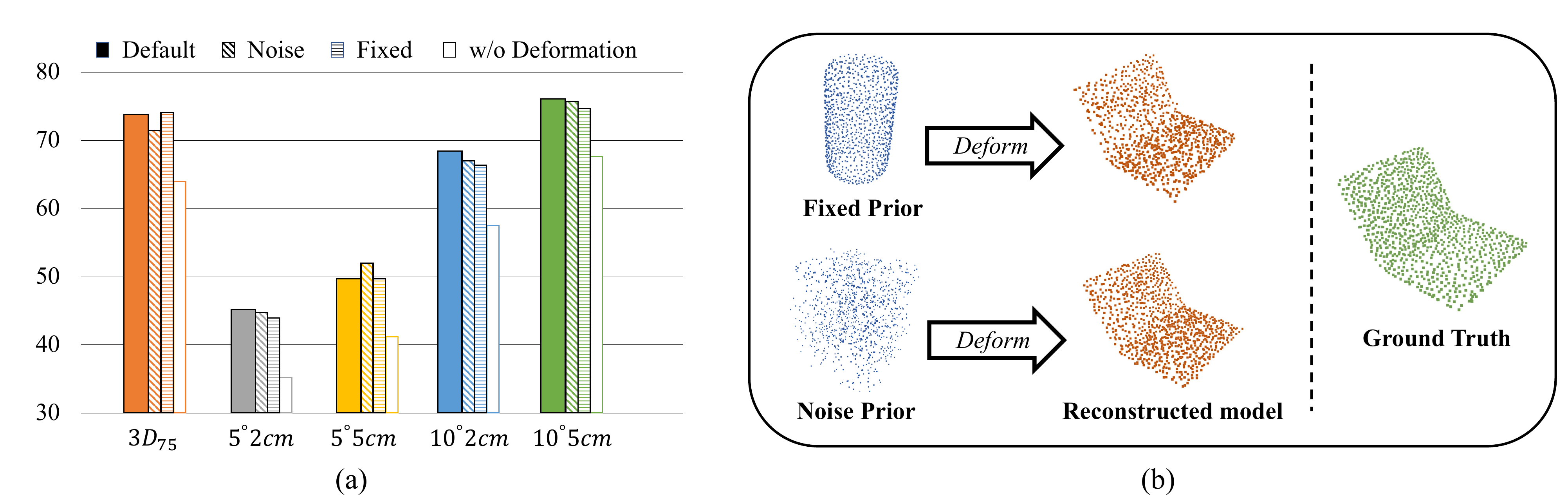}
   \caption{Illustration of the empirical experiments on the shape prior. Subfigure (a) shows the performance impact of using different forms of prior. Subfigure (b) presents the visualization of the prior deformation. Even if the class-independent and noise-generated prior are used, high-quality reconstruction results can still be obtained.} 
   \label{fig:prior-noise}
\end{center}
\vspace{-0.1cm}
\end{figure*}


This motivates us to investigate the mechanism that makes prior-based methods effective. We experiment with the shape deformation module which is used to deform the given shape prior to the desired instance(Fig. ~\ref{fig:prior-deform}) by replacing the shape priors with random noise and fixed shape prior from another category (see Fig.~\ref{fig:prior-noise}). 
We observe that the deformation module can adapt any inputs (noise or fixed prior) into a target world-space object (see Fig.~\ref{fig:prior-noise} (b)). 
Besides, the model performance remains high regardless of the 3D priors (see Fig.~\ref{fig:prior-noise}  (a)). 
The above suggests: the \textbf{shape prior itself is not necessary} for the high performance of prior-based methods, but the deformation module that learns to synthesize world-space target objects and explicitly builds the correspondence between camera and world-space is the key as the performance degrades dramatically without prior deformation. 
This promotes us to investigate new ways to build camera-to-world correspondence   without requiring 3D priors and models.

In this paper, we propose a simple yet effective prior-free model, named \textbf{I}mplicit \textbf{S}pace \textbf{T}ransformation \textbf{N}etwork (IST-Net), which implicitly sets up feature correspondence between camera-space and world-space  without requiring 3D priors or ground-truth 3D models of target objects. 
Specifically, given the camera space features, the network transforms them into world-space features which together with the camera space features are further used  for estimating camera poses. 
For learning the transformation, we propose a world-space enhancer that distills standard world-space features to supervise the transformed features. Note that the standard world-space features are obtained by transforming the input target object into its world-space with the ground-truth pose and feeding them into a feature extractor. 
Besides, given camera-space inputs, the backbone network's feature extraction capabilities are boosted by introducing an auxiliary pose estimation loss, namely the camera-space enhancer. Notably, both enhancers are training only, which brings considerable performance improvements without introducing computational overhead.  

Our main contributions are summarized as follows:

\begin{itemize}
    \item We investigate  prior-based methods and find that shape priors are not necessary for obtaining high performance while building the camera and world-space correspondence with prior deformation is a key factor.
    
    \item We propose a simple yet effective  \textbf{I}mplicit \textbf{S}pace \textbf{T}ransformation \textbf{N}etwork (IST-Net) that implicitly builds the correspondence between camera- and world-space on feature-level without requiring 3D models or priors.  
    
    \item We introduce two different space enhancers to facilitate learning the transformation and enhance their representation capability for pose estimation. 
    
    \item  We conduct a series of experiments on REAL275~\cite{wang2019normalized} and Wild6D~\cite{fu2022category} datasets to demonstrate the effectiveness of the proposed method. Notably, IST-Net is currently the only prior-free method that achieves state-of-the-art performance on the REAL275 benchmark and attains notable gains over the prior-based method in terms of efficiency and accuracy (see Fig. \ref{fig:method-compare}).

\end{itemize}

\section{Related Works}

\subsection{Prior-free Methods}
Prior-free methods focus on designing architectures for predicting the object  pose  in a concise manner. Sahin {\em et al.}~\cite{sahin2018category} propose an ``Intrinsic Structure Adaptor" to adapt the distribution shifts arising from shape discrepancies. Wang {\em et al.}~\cite{wang2019normalized} first introduce a new category-level benchmark by normalizing all object instances into a shared canonical representation named \textit{\textbf{Normalized Object Coordinate Space-(NOCS)}} and try to recover the angle of view in \textit{NOCS} for pose estimation. Chen \textit{et al.}~\cite{DBLP:journals/corr/abs-2001-09322} introduce a learned canonical shape space to handle intra-class variation. Chen \textit{et al.}~\cite{DBLP:journals/corr/abs-2008-08145} attempt to synthesize image matches upon neural rendering in order to verify the probability of each
possible pose candidate for pose estimation. Wang \textit{et al.}~\cite{DBLP:journals/corr/abs-1910-10750} propose 6D-PACK which learns to compactly represent an object by a handful of 3D
key points based on the motion information and compute the pose by tracking. In pursuit of more efficient and direct pose estimation, a few methods~\cite{fs-net, GPV-pose, dualpose-net} work on designing the network in an end-to-end manner. Chen \textit{et al.}~\cite{fs-net} decouple the rotation into two mutually orthogonal vectors to fully decode the orientation information which allows the network to naturally handle the circle symmetry object. Di \textit{et al.}~\cite{GPV-pose} embody the geometric insights with bounding box projection to enhance the learning of category-level pose-sensitive features. Lin \textit{et al.}~\cite{dualpose-net} introduce DualPoseNet which is composed of two parallel pose decoders on top of a shared pose encoder. The two decoders work in an implicit and explicit manner with the restriction of the predicted pose consistency.

\subsection{Prior-based Methods}
Since the severe intra-class variation, the generalization of the prior-free models is 
greatly suppressed. To alleviate this 
issue, some literature~\cite{fu2022category, tian2020shape, dpdn, zhang2022self, zhang2022self, weng2021captra} turn to focus on prior-based methods. Tian \textit{et al.}~\cite{tian2020shape} present a general solution. They set up shape priors for each category upon an autoencoder and then use these priors as the standard template to reconstruct the canonical model for each instance. Chen \textit{et al.}~\cite{DBLP:journals/corr/abs-2001-09322} use a variational autoencoder (VAE)~\cite{kusner2017grammar} for reconstructing standard object shape, followed by a fully sparse convolution network for pose regression. Wang \textit{et al.}~\cite{CR-Net} propose a cascaded relation network to capture the underlying relations of multi-source inputs. Kai \textit{et al.}~\cite{sgpa} utilize a transformer network~\cite{dosovitskiy2020image} to model the global structure similarity between prior and target object, based
on which the object semantic information is injected into the
prior feature to dynamically adapt the category-level prior
to each particular object. Fan \textit{et al.}~\cite{acr-pose} adopt a shape prior guided reconstruction network and a discriminator network to learn high-quality canonical representations.
Zhang \textit{et al.}~\cite{zhang2022rbp} use the shape priors as the indicator to predict pose and zero-mean residual vectors  which  encapsulate the spatial cues of the pose and enable geometry-guided consistency terms. Zhang \textit{et al.}~\cite{zhang2022self}  learn dense correspondences between input images and the canonical shape prior via surface embedding. Lin \textit{et al.}~\cite{dpdn} establish deep correspondence in the feature space between shape prior and canonical model, which yields a surprising performance boost. 


\section{Analysis of Shape Priors}

\subsection{Preliminary}

To overcome intra-class variation, the prior deformation, as a practical module, has been widely adopted by recent works ~\cite{acr-pose,sgpa, dpdn}. The vanilla version of prior deformation can be divided into two parts: 1) generating shape priors and 2) leveraging shape priors to develop prior deformation techniques. 

For the former one, the common solution is to train
an autoencoder with \textbf{various object models} sampled from
ShapeNet~\cite{chang2015shapenet}, then acquire the category-level shape em-
bedding by averaging the latent vectors output by the en-
coder. These shape embeddings will be fed into the decoder to get the shape priors. It is worth mentioning that this process needs to rely on a large number of 3D models to obtain a general prior of a category. 

\begin{figure}[t]
\begin{center}
   \includegraphics[width=1.0\linewidth]{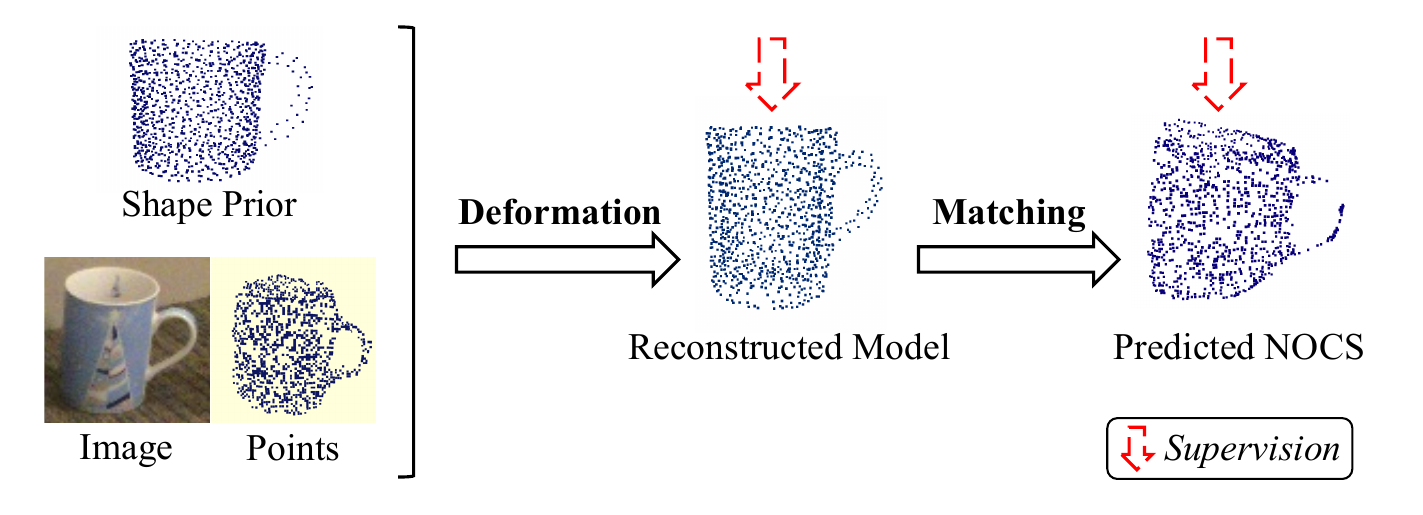}
   \caption{Illustration of the vanilla version of prior deformation. NOCS is equivalent to the coordinate in the world-space.} 
   \label{fig:prior-deform}
\end{center}
\end{figure}

For the latter one, we use Fig.~\ref{fig:prior-deform} to illustrate the process. Given the shape prior, image patch, and observed points, the network first learns a deformation field that deforms the shape prior to the desired object instance, which is \textbf{supervised by a ground-truth 3D model}. Furthermore, the network outputs a matching matrix that indicates the point-to-point correspondences between the observed target points and the reconstructed models. 
These correspondences transform the models to the viewpoint in the world coordinate system.
With the information from camera-space (depth images) and world-space (matched priors), pose parameters  can be easily solved via Umeyama algorithm~\cite{Umeyama} or pose regression by neural networks.

\subsection{Is Shape Prior Necessary?}
\label{sec:prior-not-im}

We conduct extensive experiments to verify whether the shape prior is necessary to address the intra-class variation problem. Specifically, we choose a competitive candidate from prior-based methods, DPDN~\cite{dpdn}. (For other methods please refer to the Appendix). We set up the following settings:

\begin{itemize}
    \item \textit{\textbf{Case-1:} Official baseline.}
    \item \textit{\textbf{Case-2:} All the categories share the same prior (can) in replace of the class-specific priors.}
    \item \textit{\textbf{Case-3:} Using random noise restricted to the unit cube instead of standard shape priors.}
    \item \textit{\textbf{Case-4:} Removing the prior deformation from original framework.}
\end{itemize}

\begin{figure*}[t]
\begin{center}
   \includegraphics[width=1.0\linewidth]{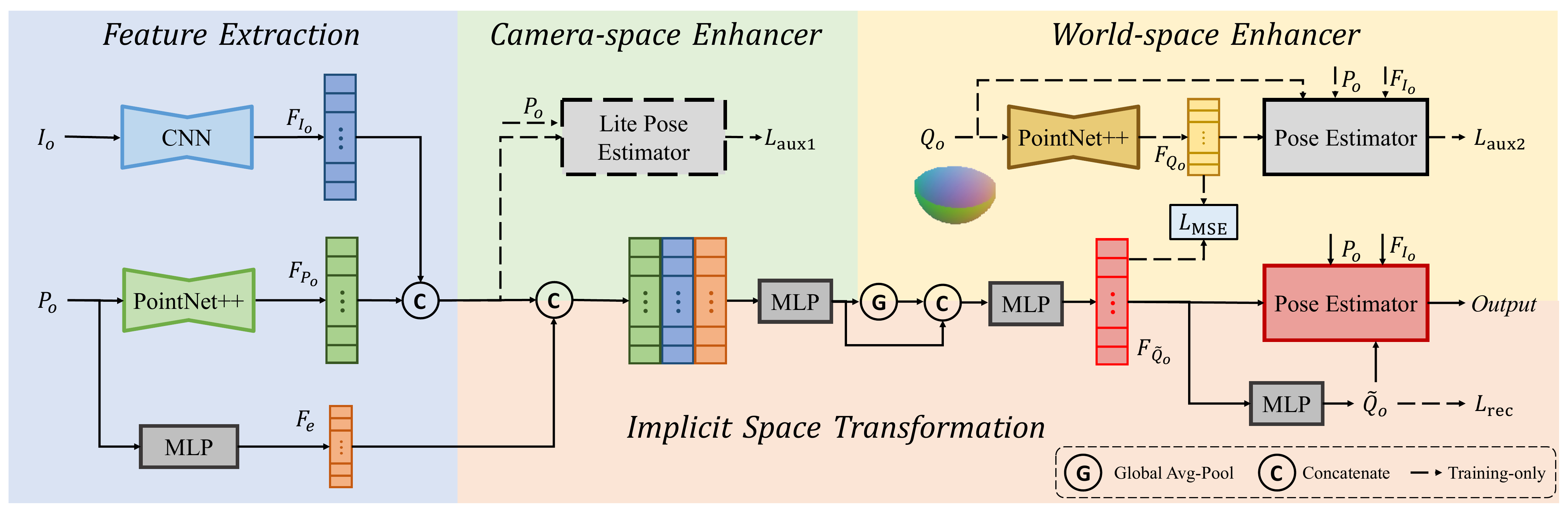}
   \caption{Illustration of our proposed Implicit Space Transformation Network
IST-Net for 6D pose estimation.} 
   \label{fig:IST-Net}
\end{center}
\end{figure*}
From Fig.~\ref{fig:prior-noise}, we can conclude that \textit{category-specific priors are not necessary} as the model can learn to deform a shape to match the target from even random noise.
The reason stems from the explicit supervision of ground-truth 3D models during training, which enables the model to learn to deform any given prior (\eg, random noise) to  reconstruct a target object.  
Nevertheless, the prior deformation module contributes to the prior-based model as the performance drops dramatically without this module (see Fig. \ref{fig:prior-noise} (a) w/o deformation). The above suggests that the key to the success of \textit{prior-based methods is the deformation module that aligns objects in the camera and world-space and facilitates building correspondences, but not the prior itself}. 


Hence, we explore new ways to transform camera space inputs to world-space and implicitly builds the correspondence between them without relying on priors or 3D ground-truth models in training. 
We present a prior-free implicit feature transformation network (\textbf{IST-Net}) with details unfolded in Sec.~\ref{sec:IST}.
On one hand, our method gets rid of the dependence on a large number of 3D models required by prior-based methods. On the other hand, we make the model aware of the information in world-space with a simple design which further allows us to develop an efficient prior-free pose estimator without sacrificing model performance.

\section{Method}

Before delving into the details, we first show some mathematical notations in the category-level 6D pose estimation task for clarity. $P_o\in\mathbb{R}^{N_o\times3}$ and $I_o\in\mathbb{R}^{h\times w \times 3}$ refer to the observed point cloud and corresponding RGB image, where $(h, w)$ denotes the size of the image and $N_o$ is the number of object points. Given these inputs, the objective is to estimate the pose of the input instance, including rotation $R \in SO(3)$, translation $t \in \mathbb{R}^3$ and size $s \in \mathbb{R}^3$.

\subsection{Overview}
\label{sec:overview}
In this section, we will introduce the detailed architecture of the proposed IST-Net, as shown in Fig.~\ref{fig:IST-Net}.
Given camera-space RGB image $I_o$ and point cloud $P_o$ as inputs, a CNN backbone network, and PointNet++ are used to extract image feature $F_{I_o}$ and point feature $F_{P_o}$ and $F_e$ respectively. 
Then, the learned features are fed into the implicit space transformation module   (Sec.~\ref{sec:IST}) which performs implicit camera-space  to world-space feature $F_{\tilde{Q}_o}$ transformation. 
We also provide a detailed analysis of our implicit design and compare it with its explicit counterparts in Sec.~\ref{sec:IST}. 
Further, given point cloud $P_o$, point-wise aligned world-space feature ($F_{\tilde{Q}_o}$) and camera-space feature ($F_{I_o}$ and $F_{P_o}$), the pose estimator directly regress the poses $\{R, t , s\}$. 
We adopt the pose estimator in DPDN~\cite{dpdn}. Details are included in the Appendix. 
Finally, we introduce two auxiliary modules, namely camera-space enhancer  and world-space enhancer (Sec.~\ref{sec:CE} and Sec.~\ref{sec:WE}), to boost feature representations and facilitate learning the implicit transformation network.


\subsection{Implicit Space Transformation}
\label{sec:IST}


Since we have shown in Sec.~\ref{sec:prior-not-im}, shape prior itself is not necessary, the important factor is how to transform camera-space inputs into world-space, align them and  build their correspondences. 
To address these issues, we propose an implicit space transformation module, which transforms camera-space features to world-space in an implicit manner without resorting to ground-truth 3D models during training.  

As depicted in Fig.~\ref{fig:IST-Net}, given an RGB image $I_o$ and corresponding object point cloud $P_o$, we first use the feature extractor to acquire semantic feature which sampled as $F_{I_o} \in \mathbb{R}^{N_o\times d}$ and geometry feature $F_{P_o} \in \mathbb{R}^{N_o\times d}$. Notably, before entering PointNet++, the point cloud will be pre-processed, which means subtracting the average value of all coordinates. In other words, the feature extractor can focus on the relative geometric relationship, avoiding the interference of  spatial locations. However, this will make the network insensitive to location information, which is indispensable for the estimation of the translation and rotation. Hence, we utilize a Multi-Layer Perceptron (MLP) to encode the accurate position information into latent space as $F_e$. Then we concatenate three of them as the input of the implicit space transformation module. Sequentially, an MLP is employed to fuse 
these input features
followed by a global average pooling layer. Then the local and global features are concatenated and fed into an MLP to get the world-space feature $F_{\tilde{Q}_o}$. This process can be formulated as follows:
\begin{align}
\begin{aligned}
&F_L \;\, = \; \text{MLP}\left(\left[F_{P_o} ,F_{I_o}, F_e\right]\right), \\
&F_G \;\, = \; G\left(F_L\right), \\
&F_{\tilde{Q}_o} = \; \text{MLP}\left( \left[ F_L, F_G\right] \right),
\end{aligned}
\end{align}
where $[\dots]$ refers to concatenation and $G$ denotes the global average pooling. $F_L$ and $F_G$ refers to the local and global feature.

It should be noted that the implicit transformation network may not certainly transform the camera space feature to world space without meaningful supervision. We thus introduce a reconstruction-based loss $L_\text{rec}$ to regularize the learning. Specifically, given $F_{\tilde{Q}_o}$ as inputs, we use another MLP to predict  reconstruct per-point coordinate in the world-space $\tilde{Q}_o$ and supervise it using ground-truth world-space coordinate $Q_0$. Note that $Q_o$ can be obtained by transforming input point cloud $P_o$ using ground-truth poses without resorting to 3D models.   
Following~\cite{tian2020shape}, we adopt smooth-L1 loss as:
\begin{equation}
\label{eq:recons}
    L_\text{rec} = L_\text{SL1}\left(\tilde{Q}_o, Q_o \right), Q_o = \Gamma (P_o, R, t, s),
\end{equation}
where $L_\text{SL1}$ denotes the Smooth-L1 loss~\cite{liu2021adaptive} and $\Gamma$ indicates the 3D geometric transformation operation according to pose. This supervision will encourage the transformed features to be in the world-space.  






\paragraph{Explicit Counterpart}
\label{sec:explicit}

\begin{figure}[t]
\begin{center}
   \includegraphics[width=1.0\linewidth]{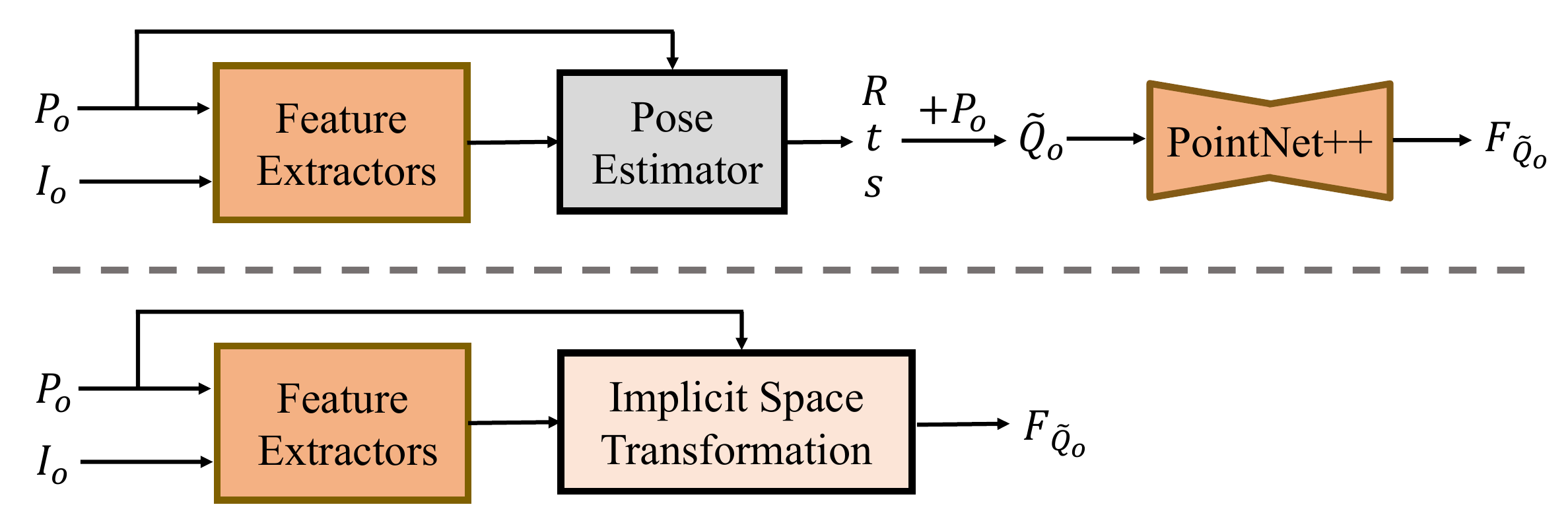}
   \caption{Comparison between implicit transformation (sub-graph below) and explicit transform(sub-graph above).} 
   \label{fig:compare-with-explicit}
\end{center}
\vspace{-0.3cm}
\end{figure}

To better showcase the merits of our implicit space transformation, we set up a comparison with its explicit counterpart. As shown in Fig.~\ref{fig:compare-with-explicit}, we firstly use an extra pose estimator to predict a group of pose parameters and then use them to explicitly transform the camera coordinate to the world coordinate as $\tilde{Q}_o$. And then we utilize another PointNet++ to extract world-space features $F_{\tilde{Q}_o}$. Finally, we feed $F_{\tilde{Q}_o}$, $\tilde{Q}_o$, and camera space items into the  estimator for pose regression. Explicit transformation is more intuitive but struggles in model redundancy. In contrast, our method is quite concise during inference, reaching a better balance between efficiency and performance. More analysis are presented in Sec.~\ref{sec:ablation}.


\subsection{Camera-space Enhancer}
\label{sec:CE}

Since the transformed feature $F_{\tilde{Q}_0}$ is derived from camera-space features, the representative power of $F_{P_0}$ and $F_{I_o}$ becomes crucial. 
Therefore, we propose a camera-space enhancement strategy to strengthen the camera-space features supervised by the auxiliary pose estimation task. The camera-space enhancer can easily be an auxiliary pose estimator whose training doesn't rely on the world-space 3D model.  This encourages the feature extractor to learn representations that bridge camera and world-space and implicitly establish a correspondence with world-space features, which is helpful for subsequent transformations and final predictions.

\subsection{World-space Enhancer}
\label{sec:WE}

In the implicit space transformation module, we predict the world-space coordinates, which encourage the network to transform camera-space features into world-space feature $F_{\tilde{Q}_o}$ and indirectly regularize the learning. Here, we propose world-space enhancer which directly enforces $F_{\tilde{Q}_o}$ to be similar to world-space counterparts   

As shown in Fig.~\ref{fig:IST-Net}, the world-space enhancer takes the world-space point cloud $Q_o$ derived using Eq.~\eqref{eq:recons} as input and produces (ground-truth) world-space features $F_{Q_o}$ using PointNet++.  $F_{Q_o}$ is used to directly supervise $F_{\tilde{Q}_o}$ as 

\begin{align}
\begin{aligned}
L_\text{feat} &= L_\text{MSE}\left(F_{\tilde{Q}_o}, F_{Q_o} \right) ~\,, \\
L_\text{MSE}\left(F_{\tilde{Q}_o}, F_{Q_o} \right) &= \frac{1}{N_o\cdot d} \sum_{i=1}^{N_o}\sum_{j=1}^{d}\Vert f_{{\tilde{Q}_o}}^{ij} - f_{Q_o}^{ij} \Vert^2 .
\end{aligned}
\end{align}

\begin{table*}[t]
\centering 
\resizebox{\linewidth}{!}{
\begin{tabular}{l | c |c c c| c  c c c c | c}
\Xhline{1.5 pt}
Method & Prior & 3$D_{25}$ & 3$D_{50}$ & 3$D_{75}$ & 5$^{\circ}2cm$ & 5$^{\circ}5cm$ & 10$^{\circ}2cm$ & 10$^{\circ}5cm$ & 10$^{\circ}10cm$ & Speed (FPS) \\
\Xhline{0.5 pt}
NOCS~\cite{wang2019normalized} &  \xmark  & \textbf{84.9} & 80.5 & 30.1 & 7.2 & 10 & 13.8 & 25.2 & 26.7 &  4.8\\

CASS~\cite{DBLP:journals/corr/abs-2001-09322} & \xmark & 84.2 & 77.7 & - & - & 23.5 & - & 58.0 & 58.3 & - \\

FS-Net~\cite{fs-net} & \xmark & - & - & - & - & 28.2 & - & 60.8 & - & 20.8\\
DualPoseNet~\cite{dualpose-net} & \xmark & - & 79.8 & 62.2 & 29.3 & 35.9 & 66.8 & - & - & 2.3\\
GPV-Pose~\cite{GPV-pose} &  \xmark & 84.2 & 83.0 & 64.4 & 32.0 & 42.9 & - & 73.3 & 74.6 & 23.6\\
\Xhline{0.5 pt}

SPD~\cite{tian2020shape} & \cmark & 83.4 & 77.3 & 53.2 & 19.3 & 21.4 & 43.2 & 54.1 & - & 6.3\\

CR-Net~\cite{CR-Net} & \cmark &- & 79.3 & 55.9 & 27.8 & 34.3 & 47.2 & 60.8 & - & -\\

SGPA~\cite{sgpa} & \cmark & - & 80.1 & 61.9 & 35.9 & 39.6 & 61.3 & 70.7 & - & 5.2 \\

DO-Net~\cite{SAR-Net} & \cmark & & 80.4 & 63.7 & 24.1 & 34.8 & 45.3 & 67.4 & - & 11.3\\
RBP-Pose~\cite{zhang2022rbp} & \cmark & - & - & 67.8 & 38.2 & 48.1 & 63.1 & 79.2 & - & 26.8\\
DPDN~\cite{dpdn} & \cmark & 84.3 & \textbf{83.4} & 76.0 & 46.0 & 50.7 & 70.4 & 78.4 & 80.4 & 25.2\\

\Xhline{0.5 pt}
IST-Net (Ours) & \xmark & 84.3 & 82.5 & \textbf{76.6}  & \textbf{47.5} & \textbf{53.4} & \textbf{72.1} & \textbf{80.5} & \textbf{82.6} & \textbf{34.6}\\

\Xhline{1.5 pt}
\end{tabular}}
\vspace{0.1cm}
\caption{Comparisons with state-of-art methods on REAL275 dataset. We summarize the pose estimation results reported in the original papers. \textbf{Prior} refers to whether the method builds upon shape priors.  `-' denotes no results are reported under this metric. All the experiments for inference speed  are conducted with a single RTX3090Ti GPU.}
\label{tab:main-results}
\end{table*}

To train the feature extractor ({\ie} PointNet++) and obtain high-quality features $F_{Q_o}$, we use another auxiliary pose estimator with the same architecture as the one that generates pose estimates to produce supervisory signals.  
Notably, the only difference from the main pose estimator (red boxes in Fig.~\ref{fig:IST-Net}) is the world-space inputs, fully encoded ground-truth world-space coordinates $Q_o$. In this way, we believe that $F_{Q_o}$ shares the same feature space with $F_{\tilde{Q}_o}$ but is more accurate, compact, and friendly for pose estimation.  

In conclusion, this module generates a standard latent feature that further provides restriction from a high-level aspect, regularizing $F_{\tilde{Q}_o}$ with implicit geometric information. 
In conjunction with constraint $L_\text{rec}$ in Sec.~\ref{sec:IST}, the space transformation is guided to learn to transform camera-space features to world-space. 

\subsection{Overall Loss Function}
In summary, the overall loss function is as below:

\begin{equation}
L = L_\text{main} + L_\text{aux1} + L_\text{aux2} + \lambda_{f}L_\text{feat} +\lambda_{r} L_\text{rec}~\,,
\end{equation}
where $\lambda_{f}$ and $\lambda_{r}$ are hyper-parameters which used to balance the individual loss contributions. $L_\text{main}$, $L_\text{aux1}$, and $L_\text{aux2}$ refer to the supervision for the outputs of main pose estimator and two feature enhancers which share the same loss format as:


\begin{equation}
\begin{aligned}
L_\text{pose} = \Vert R_\text{pred} - R_\text{gt} \Vert_{2} & + \Vert t_\text{pred} - t_\text{gt} \Vert_{2} \\ & + \Vert s_\text{pred} - s_\text{gt} \Vert_{2} \,,
\end{aligned}
\end{equation}
where subscripts ``$\text{pred}$'' and ``$\text{gt}$'' denote the predicted and ground truth pose parameters, respectively.

\section{Experiments}

\subsection{Datasets}
\noindent
\textbf{REAL275 \& CAMERA25:} Our method is trained on both the virtual dataset, CAMERA25, and the real dataset, REAL275~\cite{wang2019normalized}, and conducted an evaluation on  REAL275 test split. CAMERA25 contains 300k synthetic RGB-G images, which are generated by rendering 1,085 synthetic objects with real-world backgrounds. REAL275 includes 8k RGB-D images, where 4300 images are split for training, 950 images for validation, and 2750 images for testing. In both datasets, there are 6 categories, including bottle, bowl, camera, can, laptop, and mug.

\noindent
\textbf{Wild6D:} Wild6D~\cite{fu2022category} contains 5,166 videos with 1722 object instances and 5 categories (bottle, bowl, camera, laptop, and mug). Among this data, 486 videos are split for model evaluation.

\subsection{Implementation Details}

Following previous work~\cite{dpdn}, we use the off-the-shelf MaskRCNN~\cite{he2017mask} with a backbone of ResNet101~\cite{resnet} for generating high-quality instance masks. We adopt a PSP Network~\cite{pspnet} based on ResNet-18~\cite{resnet} for semantic feature extraction and PointNet++~\cite{pointnet2} for point-level feature extraction. The number of object point $N_{o}$ is set as 1024  and the size of the RGB image is resized to $192\times192$. We adopt several commonly used data augmentation, including random uniform noise, random rotational and translational perturbations, and  bounding box-based adjustment, which is proposed by FS-Net~\cite{fs-net}. The hyper-parameters of the loss weights $\lambda_{f}$ and $\lambda_{r}$ are set to 10 and 1, respectively. All the experiments are conducted on 2 RTX3090Ti GPUs with a batch size of 24, and the ratio of real data to synthetic data is 1:3. For a fair comparison, the total training epoch is fixed to 30 epochs, and all the modules are trained in an end-to-end manner. During inference, only the feature extractor and implicit space transformation module are preserved.

\subsection{Evaluation Metrics}

We follow~\cite{wang2019normalized, GPV-pose} and utilize the widely adopted metrics for evaluation, including the mean precision of 3D intersection over union (IoU) to jointly evaluate rotation, translation, and size. Besides, the $5^{\circ}2cm$, $5^{\circ}5cm$, $10^{\circ}2cm$, $10^{\circ}5cm$ and $10^{\circ}10cm$ are used to evaluate the rotation and translation error directly, specifically, only the prediction error under both thresholds can be considered correct.

\begin{table}[t]
\centering 

\resizebox{\linewidth}{!}{
\begin{tabular}{l | l |c c| c c c c }
\Xhline{1.5 pt}
Method & Data  & 3$D_{25}$ & 3$D_{50}$ & 5$^{\circ}2cm$ & 5$^{\circ}5cm$ & 10$^{\circ}2cm$ & 10$^{\circ}5cm$ \\
\Xhline{0.5 pt}

CASS~\cite{DBLP:journals/corr/abs-2001-09322} & C+R & 19.8 & 1.0 & 0.0 & 0.0 & 0.0 & 0.0 \\

SPD~\cite{tian2020shape} & C+R & 55.5 & 32.5 & 2.6 & 3.5 & 9.7 & 13.9 \\

DualPoseNet~\cite{dualpose-net} & C+R & 90.0 & 70.0 & 17.8 & 22.8 & 26.3 & 36.5 \\

GPV-Pose~\cite{GPV-pose} & C+R & 91.3  & 67.8 & 14.1 & 21.5 & 23.8 & 41.1 \\

RePoNet~\cite{fu2022category} & C+W$^*$ &  84.7 & 70.3 & 29.5 & 34.4 & 35.0 & 42.5 \\

Self-Pose~\cite{zhang2022self} & W$^*$ & 92.3 & 68.2 & \textbf{32.7} & 35.3 & \textbf{38.3} & \textbf{45.3}\\

\Xhline{0.5 pt}
IST-Net (Ours) & C+R & \textbf{93.4} & \textbf{79.6} & 30.7 & \textbf{35.8}  & 37.1 & 43.7\\

\Xhline{1.5 pt}
\end{tabular}}
\vspace{0.1cm}
\caption{Comparison with state-of-art methods on Wild6D dataset. The “Data” column refers to the data type for training. C=CAMERA25, R=REAL275, W=Wild6D, $^*$=not using pose annotation.}
\vspace{-0.1cm}
\label{tab:main-results-wild6d}
\end{table}

\subsection{Comparison with State-of-the-Arts}

\begin{figure}[t]
\begin{center}
   \includegraphics[width=1.0\linewidth]{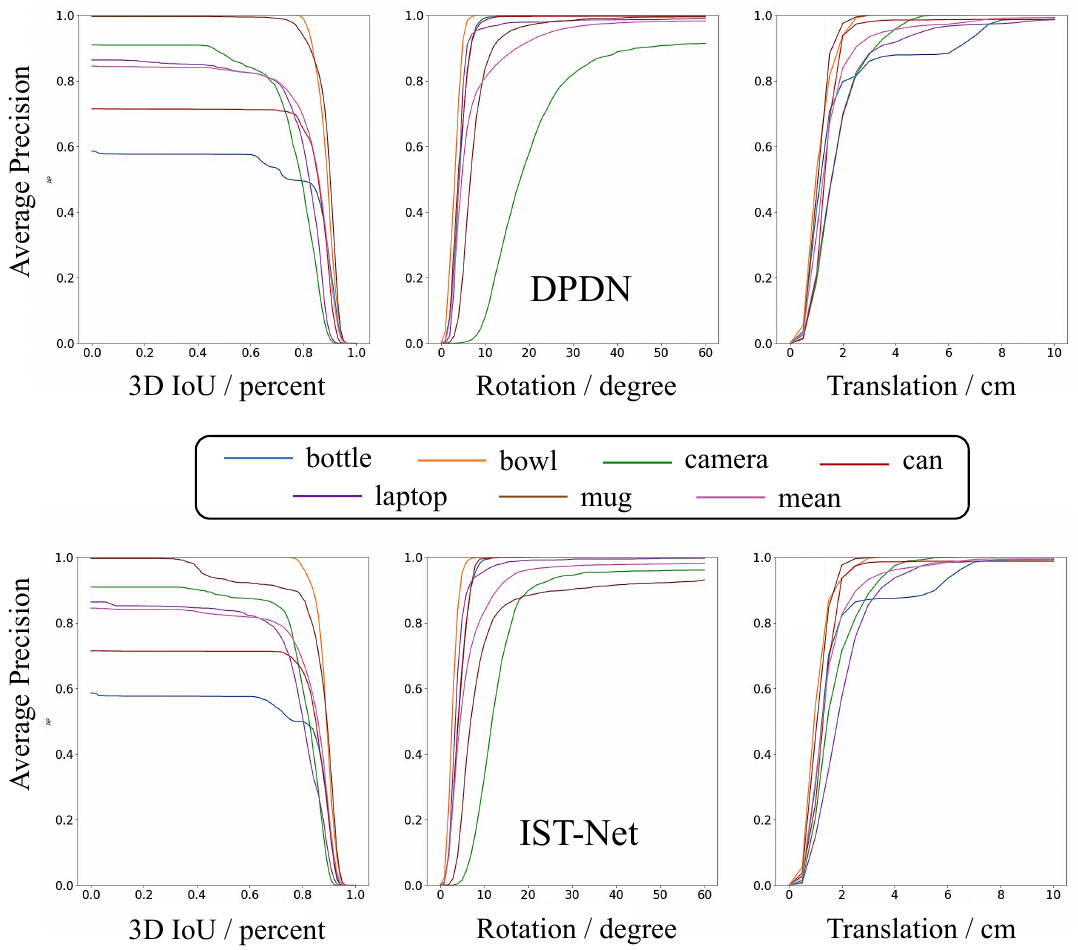}
   \caption{Quantitative comparison with DPDN~\cite{dpdn} on REAL275 in terms of average precision in 3D IoU, Rotation and Translation.} 
   \label{fig:error-curve}
\end{center}
\vspace{-0.5cm}
\end{figure}

We present the results of IST-Net with state-of-the-art methods on REAL275~\cite{wang2019normalized}, as shown in Tab.~\ref{tab:main-results}. For comparison with prior-free methods, we surpass others with a large gap on all evaluation metrics, \textit{e.g.,} we reach 47.5 and 76.6 on 5$^{\circ}2cm$ and $D_{75}$, which outperform GPV-Pose by 19.4\% and 41\%. As for prior-based methods, compared with the current most powerful method DPDN~\cite{dpdn}, we still perform significant improvements in most of the metrics. \textit{e.g.,} 76.6 vs 76 on $D_{75}$, 47.5 vs 46.0 on  5$^{\circ}2cm$, 53.4 vs 50.7 on  5$^{\circ}5cm$, 80.5 vs 78.4 on 10$^{\circ}5cm$ and 82.6 vs 80.4. Notably, it is the first time for the prior-free methods to achieve comparable or even higher performance compared with prior-based methods on REAL275.  
In addition, we present a per-class comparison between IST-Net and DPDN, as shown in Fig.~\ref{fig:error-curve}. Notably, our method performs better on the prediction of rotation, the error curve is steeper on the geometrically complex object, \textit{e.g.,} camera, which clearly proves the effectiveness of our proposed contributions. Apart from the performance, model efficiency is also worthy of attention, we list the inference time in the last column of Tab.~\ref{tab:main-results}. IST-Net reaches top inference speed which far exceeds other methods by more than \textbf{25\% acceleration}. 

We conduct experiments on a larger dataset, Wild6D~\cite{fu2022category}, to further verify the effectiveness of the proposed method.  We directly test our model which are trained on REAL275 and CAMERA25 datasets without extra fine-tuning. The results are reported in Tab~\ref{tab:main-results-wild6d}. It can be observed that IST-Net is much better than those designed for the REAL275 dataset and has an obvious improvement in various matrices. Compared with RePoNet~\cite{fu2022category} and Self-Pose~\cite{zhang2022self} both of \textit{which are trained upon Wild6D, our method shows good generalization without 
fine-tuning on the target dataset}. We achieve the highest performance of 93.4 and 79.6 on 3$D_{25}$  and 3$D_{50}$. As for other metrics, our method can also reach a similar performance to Self-Pose. These analysis and results demonstrate the potential of our method.




\subsection{Ablation Study}
\label{sec:ablation}

\noindent
\textbf{Effects of Proposed Modules.}
We ablate the combination of different modules of the proposed method, the results are shown in Tab~\ref{tab:ablate-modules}. Firstly we present the effectiveness of the implicit space transformation module (IST). By adding this module, we can easily observe that the baseline is greatly lifted, suggesting that  transforming camera-space features to world-space counterparts and building the correspondence between them in an implicit manner indeed can benefit the pose estimation. Besides this, after adding the camera-space enhancer (CE), the precision on 5$^{\circ}5cm$ increases from 48.5 to 52.5, the reason is that with this auxiliary module, the feature extractors are enriched with more pose-sensitive information, which 
is beneficial to the quality of feature transformation and to improve the accuracy of final pose estimation. In addition, we show the advantage of world-space enhancer (WE), by combining it with IST. The results (E4) show that WE can further extend the performance, especially on  10$^{\circ}2cm$ and 10$^{\circ}5cm$, which indicates that high-level supervision provides additional information different from low-level constraint. Finally, by combining all modules together, we reach relatively competitive performance.

\begin{table}[t]
\centering 
\resizebox{\linewidth}{!}{\begin{tabular}{ c | c |c | c | c  c c c c }
\Xhline{1.5 pt}
 & IST & CE & WE & 3$D_{75}$ & 5$^{\circ}2cm$ & 5$^{\circ}5cm$ & 10$^{\circ}2cm$ & 10$^{\circ}5cm$\\

\Xhline{0.5 pt}
E1 &  &  &  & 70.7 & 35.2 & 42.9 & 58.8 & 72.2 \\
E2 & \cmark & & & 72.2  & 42.9 & 48.5 & 68.7 & 78.1 \\
E3 & \cmark & \cmark &  & 73.9 & 44.2 & 52.5 & 68.5 & 79.2\\
E4 & \cmark &  & \cmark & 75.9 & 43.5 & 48.9 & 70.4 & 80.4\\
E5 & \cmark & \cmark & \cmark & \textbf{76.6}  & \textbf{47.5} & \textbf{53.4} & \textbf{72.1} & \textbf{80.5}\\

\Xhline{1.5 pt}
\end{tabular}}
\vspace{0.1cm}
\caption{Ablation on different configurations of network architectures. IST refers to the implicit space transformation, CE and WE denote the camera and world-space enhancement. Note: In E1 we just remove the supervision of IST, not the module itself.}
\vspace{-0.1cm}
\label{tab:ablate-modules}
\end{table}

\begin{table}[h]
\centering 
\resizebox{\linewidth}{!}{\begin{tabular}{ c | c c | c c c c }
\Xhline{1.5 pt}
 PE & 3$D_{50}$ & 3$D_{75}$ & 5$^{\circ}2cm$ & 5$^{\circ}5cm$ & 10$^{\circ}2cm$ & 10$^{\circ}5cm$ \\

\Xhline{0.5 pt}
w  & \textbf{82.5} & \textbf{76.6} & \textbf{47.5} & \textbf{53.4} & \textbf{72.1} & \textbf{80.5} \\
w/o & 81.3 & 73.0 & 41.5 & 47.4 & 68.8 & 78.8 \\

\Xhline{1.5 pt}
\end{tabular}}
\vspace{0.1cm}
\caption{Ablation on position encoding term. PE refers to the position encoding term.}
\label{tab:ablate-PE}
\vspace{-0.2cm}
\end{table}

\begin{table}[h]
\centering 
\resizebox{\linewidth}{!}{\begin{tabular}{ c | c | c c c c | c c }
\Xhline{1.5 pt}
 & 3$D_{75}$ & 5$^{\circ}2cm$ & 5$^{\circ}5cm$ & 10$^{\circ}2cm$ & 10$^{\circ}5cm$ & Param. & FPS\\

\Xhline{0.5 pt}
Implicit  & 73.9 & 44.2 & \textbf{52.5} & 68.5 & 79.2 & \textbf{21M} & \textbf{34} \\
Explicit  & \textbf{75.1} & \textbf{45.0} & 50.7 & \textbf{69.6} & \textbf{80.0} & 24M & 22 \\

\Xhline{1.5 pt}
\end{tabular}}
\vspace{0.1cm}
\caption{Comparison with explicit space transformation. Param. refers to the number of parameters.}
\vspace{-0.1cm}
\label{tab:ablate-explicit}
\end{table}

\begin{table}[h]
\centering 
\resizebox{\linewidth}{!}{\begin{tabular}{ c | c c | c c c c }
\Xhline{1.5 pt}
 Method & 3$D_{50}$ & 3$D_{75}$ & 5$^{\circ}2cm$ & 5$^{\circ}5cm$ & 10$^{\circ}2cm$ & 10$^{\circ}5cm$ \\

\Xhline{0.5 pt}
SGPA~\cite{sgpa}  & 80.1 & 61.9 & \textbf{35.9} & \textbf{39.6} & 61.3 & 70.7 \\
IST-Net$^M$ & \textbf{82.5} & \textbf{72.7} & 35.8 & 38.4 & \textbf{64.2} & \textbf{72.4} \\
\Xhline{1.5 pt}
\end{tabular}}
\vspace{0.1cm}
\caption{Ablation on the predicted world coordinate. $M$ is our matching-based variant, using the Umeyama algorithm for testing.}
\label{tab:ablate-predict-nocs}
\vspace{-0.1cm}
\end{table}





\vspace{0.05in}\noindent
\textbf{Effects of Position Encoding Term.}
In this part, we verify the effect of the positional encoding (PE) term, the results are shown in Tab.~\ref{tab:ablate-PE}. Obviously, without the PE module, the performance drops significantly on 5$^{\circ}2cm$ and 5$^{\circ}5cm$. This in turn  proves that the PE module makes up for the position lost by the feature extractor, which benefits pose regression.






\vspace{0.05in}\noindent
\textbf{Comparison with Explicit Space Transformation.}
To further verify the effectiveness of the proposed implicit space transformation, we set up an experiment with its explicit counterpart. For reaching a fair comparison, only WE and IST are included in the implicit candidate. From the results shown in Tab.~\ref{tab:ablate-explicit}, we can easily find the results of the two methods are very close. However, our method yields obvious superiority in speed (34Hz vs 22Hz) and parameter quantity (21M vs 24M) which attributes to the succinct feature space transformation instead of introducing repetitive modules for extracting features from coordinates. This further indicates the potential of the proposed modules.




\begin{figure}[t]
\begin{center}
   \includegraphics[width=1.0\linewidth]{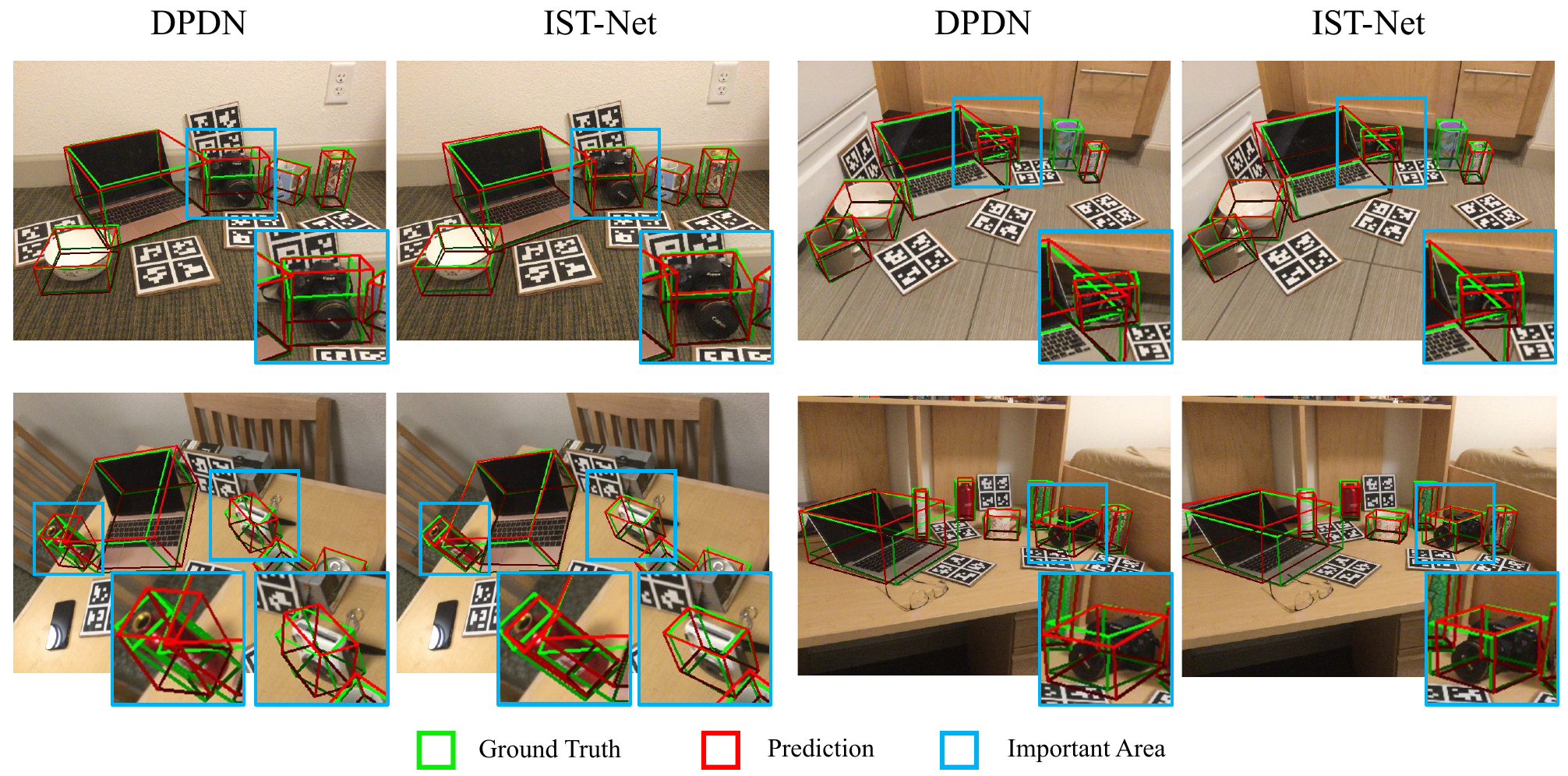}
   \caption{Qualitive comparison between IST-Net and DPDN on REAL275 dataset.} 
   \label{fig:visualization}
\end{center}
\vspace{-0.6cm}
\end{figure}

\vspace{0.05in}\noindent
\textbf{Ablations on Predicted World Coordinate.} 
Considering that we predict the coordinate of observed points in world-space for supervising implicit space transformation from a low-level perspective. Therefore, the quality of the generated coordinate can also reflect the effectiveness of the proposed method. To verify this, we use the predicted coordinate and observed points in camera space for solving the pose parameters by Umeyama algorithm~\cite{Umeyama}. As shown in Tab.~\ref{tab:ablate-predict-nocs}, our method achieves comparable results with SGPA, even with significant improvement on 3$D_{75}$ (72.7 vs 61.9), indicating that the network can reconstruct the perspective in world-space without introducing shape prior.

\section{Qualitative Analysis}
In Fig~\ref{fig:visualization}, we visually compare our methods and DPDN on the REAL275 dataset. It clearly shows the superiority of
our method. As highlighted in blue boxes, DPDN easily gets stuck in the object with complex structure, \textit{e.g.,} camera, which presents as apparent deviations of predicted boxes.This reflects the prior deformation has a poor capability for modeling challenging cases. By contrast, IST-Net demonstrates strong performance in predicting accurate rotation and translation estimations. The reason is that with the implicit transformation, the geometric structures are transmitted to the world-space together with the feature, which ensures the sensitivity of the network to complex structures.

\section{Conclusion}

In this paper, we analyze the overlooked issues in prior-based pose estimation methods and empirically find that shape prior does not contribute to performance boosts. The keypoint is actually the deformation process, which builds correspondence between camera and world coordinates by reconstructing the object shape in the world space. Inspired by this, we design an implicit space transformation network (IST-Net) to transform the camera-space features to world space in an implicit manner. It builds the space correspondence without requiring 3D priors or ground-truth 3D models of target objects. Besides, we design two independent feature enhancers to further enhance the features from both camera- and world-space, which enriches them with more pose-sensitive information and geometrical constraints.
Extensive experiments on the challenging benchmark show the effectiveness of our method in both efficiency and accuracy. 
We hope our investigation can provide new insights for future research in the community.

{\small
\bibliographystyle{ieee_fullname}
\bibliography{egbib}
}

\appendix

\newpage

\section*{Appendix}

\section{More Implementation Details}
\label{sec:impelement}

\subsection{Training and Inference Details.}

We train our IST-Net from scratch in an end-to-end manner for 30 epochs with a batch size of 24. We further employ the Adam optimizer with a base learning rate of 0.01. We adopt the StepLR scheduler with step size 1 and gamma as 5. Our experiments are conducted on two  RTX3090Ti GPUs. 

\subsection{Network Configurations}
\begin{figure}[h]
\begin{center}
   \includegraphics[width=1.0\linewidth]{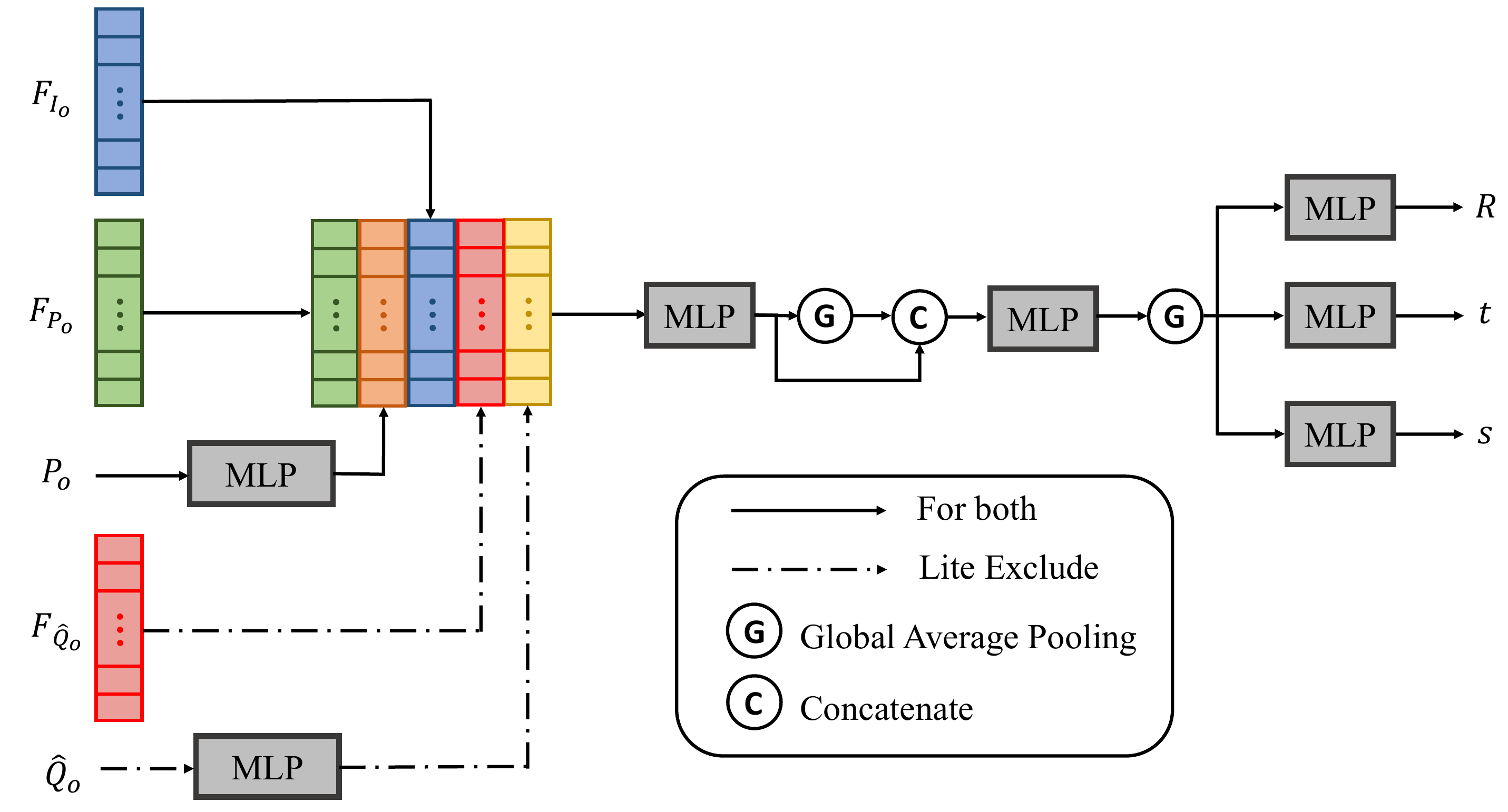}
   \caption{Architecture of pose estimators. The solid lines represent the same parts of all estimators, and the dashed line represents the part that is not adopted by the lite pose estimator.} 
   \label{fig:pose-estimator}
\end{center}
\end{figure}

As mentioned in the main paper,  we provide the detailed architecture of the pose estimators, as shown in Fig.~\ref{fig:pose-estimator}. IST-Net contains three pose estimators in camera-space enhancer, world-space enhancer, and final pose regression which follow similar architectures.
The pose estimators in world-space enhancer and final pose regression share the same architecture and adopt a standard design, namely standard pose estimator. While the pose estimator in camera-space enhancer adopts a lightweight design, namely lite pose estimator. Specifically, in Fig.~\ref{fig:pose-estimator}, the lite pose estimator only takes camera space information as input, including semantic features $F_{P_o}$, geometrical features $F_{I_o}$ and position encoding term which is generated by MLP upon $P_o$. For the standard pose estimator, its inputs contain extra information from world-space, including world-space geometrical features $F_{\hat{Q}_o}$ and world-space position encoding term. Then the inputs are concatenated together and sent into an MLP to yield the fused features
followed by a global average pooling layer. We further concatenate the global and local features and use a combination of MLP and a pooling layer to acquire the compressed features. Finally, three independent MLPs are used to predict $R$, $t$, and $s$ respectively.

\section{More Experimental Results}
\label{sec:more-results}

We further report the results of our method on the CAMERA25 dataset, as shown in Tab.~\ref{tab:main-results-syn}. Our method is competitive with other methods, specifically, on metric 3$D_{75}$, IST-Net outperforms the previous state-of-the-art method by 2\%. This indicates that our method has a strong ability to comprehensively estimate rotation, translation, and size.

\begin{table}[t]
\centering 

\resizebox{\linewidth}{!}{
\begin{tabular}{l | c |c c| c c c c }
\Xhline{1.5 pt}
Method & Prior  & 3$D_{50}$ & 3$D_{75}$ & 5$^{\circ}2cm$ & 5$^{\circ}5cm$ & 10$^{\circ}2cm$ & 10$^{\circ}5cm$ \\
\Xhline{0.5 pt}

NOCS~\cite{wang2019normalized} &  \xmark  & 83.9 & 69.5 & 32.3 & 40.9 & 48.2  & 64.6  \\

DualPoseNet~\cite{dualpose-net} & \xmark & 92.4 & 86.4 & 64.7 & 70.7 & 77.2 & 84.7 \\
GPV-Pose~\cite{GPV-pose} &  \xmark & 93.4 & 88.3 & 72.1 & 79.1 &  - & 89.0 \\
\Xhline{0.5 pt}

SPD~\cite{tian2020shape} & \cmark & 93.2 & 83.1 & 54.3 & 59.0  & 73.3 & 81.5\\

CR-Net~\cite{CR-Net} & \cmark & \textbf{93.8} & 88.0 & 72.0 & 76.4 & 81.0  & 87.7\\

SGPA~\cite{sgpa} & \cmark & 93.2 & 88.1 & 70.7 & 74.5 &  \textbf{82.7} & 88.4\\

RBP-Pose~\cite{zhang2022rbp} & \cmark & 93.1 & 89.0 & \textbf{73.5}  & 79.6 & 82.1 & 89.5\\

\Xhline{0.5 pt}
IST-Net (Ours) & \xmark & 93.7 & \textbf{90.8} & 71.3 & \textbf{79.9}  & 79.4 & \textbf{89.9}\\

\Xhline{1.5 pt}
\end{tabular}}
\vspace{0.1cm}
\caption{Comparison with state-of-art methods on CAMERA25 dataset. We summarize the pose estimation results reported in the original papers. \textbf{Prior} refers to whether the method builds upon shape priors.  `-' denotes no results reported under this metric.}
\label{tab:main-results-syn}
\end{table}

\section{More Ablation Studies}
\label{sec:more-ablation}

\subsection{Ablate on Loss Weight}

We further ablate the effect of different choices of $\lambda_f$ on pose accuracy. We gradually enlarge the $\lambda_f$ from 1 to 100. The comparative results are shown in Tab.~\ref{tab:ablate-loss-weight}. When $\lambda_f$ is too small, the supervision is limited, and when it is too large, the supervision from ground truth will be weakened. Overall, when $\lambda_f$ is set as 10, we reach the best performance.

\begin{table}[t]
\centering 
\resizebox{\linewidth}{!}{\begin{tabular}{ c | c c | c c c c c}
\Xhline{1.5 pt}
 $\lambda_f$ & 3$D_{50}$ & 3$D_{75}$ & 5$^{\circ}2cm$ & 5$^{\circ}5cm$ & 10$^{\circ}2cm$ & 10$^{\circ}10cm$ & 10$^{\circ}5cm$ \\

\Xhline{0.5 pt}
1 & 81.6 & 72.6 & 40.8 & 47.0 & 68.0 & 77.7 & 79.8 \\
3 & 82.7 & 76.1 & 42.9 & 48.9 & 70.1 & 80.1 & 82.1 \\
5 & \textbf{83.2} & 76.1 & 43.6 & 50.6 & 69.1 & 79.7 & 81.8 \\
10 & 82.5 & \textbf{76.6} & \textbf{47.5} & \textbf{53.4} & \textbf{72.1} & \textbf{80.5} & \textbf{82.6}\\
20 & 82.0 & 75.2 & 45.0 & 51.5 & 68.0 & 77.9 & 80.0 \\
50  & 82.8 & 75.3 & 41.5 & 47.2 & 68.8 & 77.9 & 80.0\\
100 & 83.1 & 76.3  & 45.3 & 50.4 & 70.4 & 78.9 & 81.1 \\

\Xhline{1.5 pt}
\end{tabular}}
\vspace{0.1cm}
\caption{Ablate on the loss weight $\lambda_f$.}
\label{tab:ablate-loss-weight}
\end{table}

\begin{figure*}[t]
\begin{center}
   \includegraphics[width=0.7\linewidth]{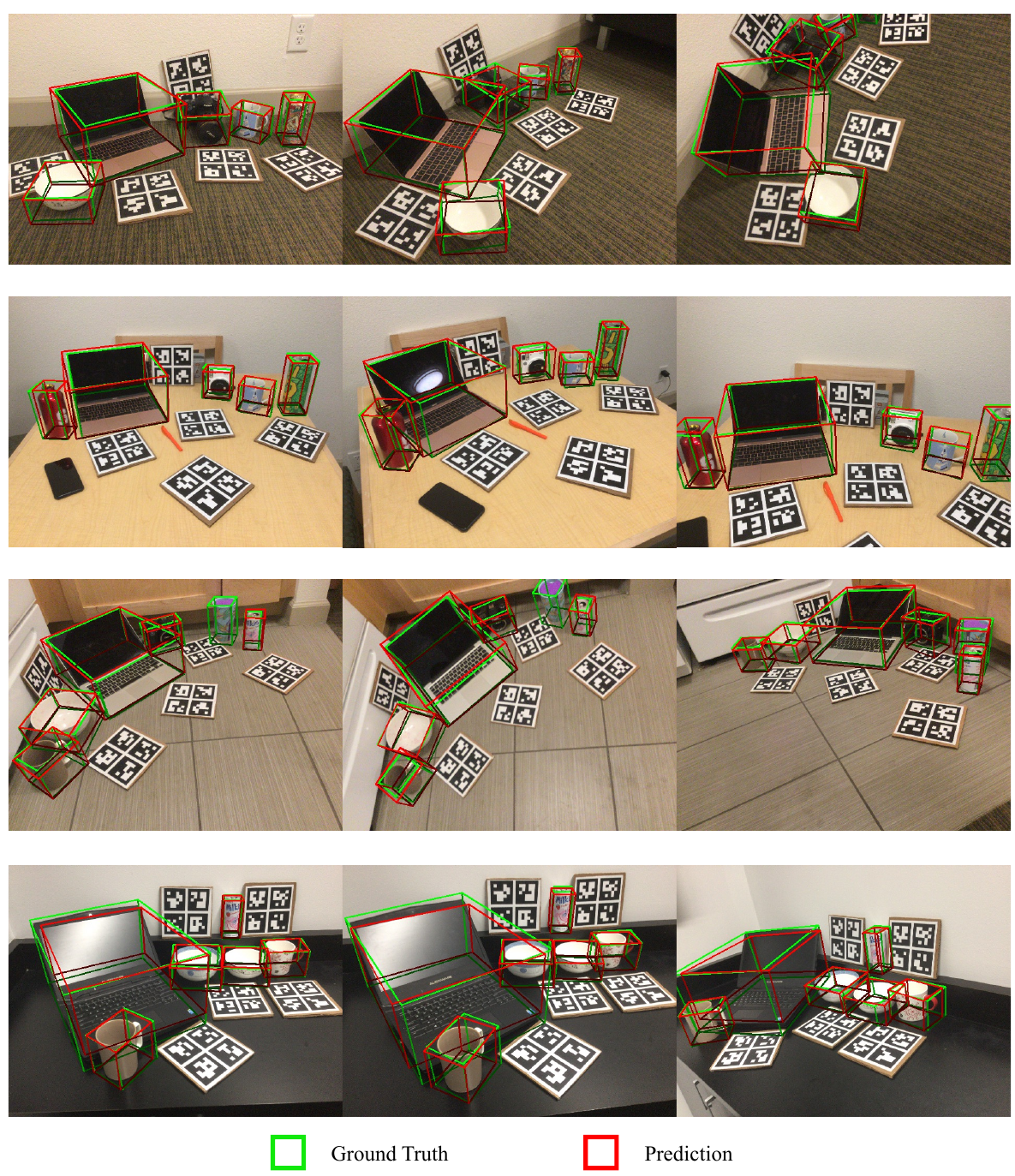}
   \caption{More visualization on REAL275 dataset.} 
   \label{fig:more-vis}
\end{center}
\vspace{-0.3cm}
\end{figure*}

\subsection{Ablate on Loss Type}

In this experiment, we ablate the effect of different loss types of $L_\text{feat}$. We present the results of MSE loss and L1 loss in 
Tab.~\ref{tab:ablate-loss-type}. In contrast, MSE Loss has more advantages. It does not require two features to be completely similar but imposes strong constraints on places with large differences, which makes the imitation between features easier to learn.

\begin{table}[t]
\centering 
\resizebox{\linewidth}{!}{\begin{tabular}{ c | c c | c c c c c}
\Xhline{1.5 pt}
Type & 3$D_{50}$ & 3$D_{75}$ & 5$^{\circ}2cm$ & 5$^{\circ}5cm$ & 10$^{\circ}2cm$ & 10$^{\circ}5cm$ & 10$^{\circ}10cm$ \\

\Xhline{0.5 pt}

MSE & 82.5 & \textbf{76.6} & \textbf{47.5} & \textbf{53.4} & \textbf{72.1} & \textbf{80.5} & \textbf{82.6} \\
L1 & \textbf{82.8} & 74.6 & 44.6 & 50.4 & 69.8 & 78.8 & 81.1\\

\Xhline{1.5 pt}
\end{tabular}}
\vspace{0.0cm}
\caption{Ablate on the loss type of $L_{feat}$}
\label{tab:ablate-loss-type}
\end{table}

\subsection{Ablate on Shape Priors with Different Methods}

In this part, we provide more experimental results to support the assumption ``shape priors are not necessary'' which is detailed in the main paper. We choose two competitive candidates from matching-based and regression-based methods, DPDN~\cite{dpdn} and SGPA~\cite{sgpa}, using prior deformation. We list the experimental results in Tab.~\ref{tab:ablate-shape-prior}. We can find that regardless of whether the approach is a matching-based or a direct regression-based method when we use category-independent prior and noise  to replace the default shape prior, the final performance does not have a significant difference. This phenomenon further reflects that shape prior is redundant for the prior deformation process, supporting our major claims in the main paper.

\begin{table}[t]
\centering 
\resizebox{\linewidth}{!}{
\begin{tabular}{l | l |c c c| c c c c c}
\Xhline{1.5 pt}
Method & Prior & 3$D_{25}$ & 3$D_{50}$ & 3$D_{75}$ & 5$^{\circ}2cm$ & 5$^{\circ}5cm$ & 10$^{\circ}2cm$ & 10$^{\circ}5cm$ & 10$^{\circ}10cm$ \\
\Xhline{0.5 pt}
\multirow{8}*{SGPA~\cite{sgpa}} & \textcolor{blue}{default}  & \textcolor{blue}{-} & \textcolor{blue}{80.1} & \textcolor{blue}{61.9} & \textcolor{blue}{35.9} & \textcolor{blue}{39.6} & \textcolor{blue}{61.3} & \textcolor{blue}{70.7} & \textcolor{blue}{-} \\

 &  bottle & 83.9 & 81.0 & 65.5 & 37.0 & 42.1 & 58.6 & 69.9 & - \\
 & bowl  & 84.0 & 81.2 & 64.3 & 36.2 & 40.7 & 60.5 & 70.9 & -  \\
 &  camera & 83.8 & 79.9 & 62.6 & 35.4 & 39.7 & 59.5 & 69.9 &  - \\
  &  can & 84.1 & 80.8 & 65.1 & 36.5 & 41.5 & 59.3 & 70.4 & - \\
   &  laptop & 83.7 & 79.2 & 63.5 & 38.7 & 42.7 & 61.0 & 71.6 & - \\
    &  mug & 83.8 & 80.1 & 64.1 & 35.0 & 40.1 & 59.7 &  68.2 & - \\
    &  noise & 83.8 & 79.9 & 60.3 & 35.2 & 39.6 & 59.5 & 69.7 & - \\
\Xhline{0.5 pt}
\multirow{8}*{DPDNs~\cite{dpdn}} &  \textcolor{blue}{default} & \textcolor{blue}{84.2} & \textcolor{blue}{83.4} & \textcolor{blue}{76.0} & \textcolor{blue}{46.0} & \textcolor{blue}{50.7} & \textcolor{blue}{70.4} & \textcolor{blue}{78.4} & \textcolor{blue}{80.4} \\

 &  bottle & 84.0 & 83.3 & 74.6 & 46.2 & 50.4 & 67.5 & 77.2 & 79.2 \\
 & bowl  & 83.8 & 83.2 & 75.9 & 46.1 & 51.3 & 68.0 & 78.1 & 80.1  \\
 &  camera & 84.0 & 82.3 & 73.5 & 45.5 & 53.1 & 66.9 & 77.9 & 80.1 \\
  &  can &  84.2 &  83.9 & 76.3 & 44.6 & 50.7 & 68.2 & 77.0 & 79.3 \\
   &  laptop & 83.4 & 81.4 & 73.2 & 44.2 & 49.2 & 67.9 & 77.2 & 79.9 \\
    &  mug & 84.1 & 84.0 & 76.6 & 45.9 & 50.3 & 68.9 & 77.4 & 79.7 \\
    &  noise & 84.2 & 83.8 & 76.1 & 45.7 & 51.0 & 69.5 & 77.7 & 79.8 \\


\Xhline{1.5 pt}
\end{tabular}}
\vspace{0.1cm}
\caption{Ablate on shape priors with different Methods. \textcolor{blue}{"default"} represents the standard result obtained from the original paper.  ‘-’ denotes no results are reported in the original literature. }
\label{tab:ablate-shape-prior}
\end{table}

\section{More Visualization}
\label{sec:more-vis}

As shown in Fig.~\ref{fig:more-vis}, we show more visualization of IST-Net on the REAL275 test split. As
highlighted with the red box, ours can accurately
predict the object pose,  which visually demonstrates the superiority of our method.

\section{Limitation Analysis and Future Work}
\label{sec:limitation} 

Our method  yields strong performance in NOCS and Wild6D datasets, but it might be sufficient for in-the-wild open-world evaluation, because, existing datasets contain limited object categories and the object structure is relatively simple.

We will work on building a category-level dataset with deiverse object types and shapes to further push forward the area. We hope our current investigation can shed light on more new insights in pose estimation.

\end{document}